\documentclass[dvipsnames,format=sigconf,preprint=true,review=false]{acmart}

\AtBeginDocument{%
  }
\usepackage{graphicx}
\usepackage{booktabs}
\usepackage{algorithm}
\usepackage{algorithmic}
\usepackage{enumitem}
\usepackage{subcaption}
\usepackage{stfloats}

\begin{document}

\title{Continuous Program Search}


\author{Matthew Siper}
\affiliation{%
  \institution{Nof1}
  \city{New York}
  \state{NY}
  \country{USA}
}
\email{m@thenof1.com}

\author{Muhammad Umair Nasir}
\affiliation{%
  \institution{Nof1}
  \city{New York}
  \state{NY}
  \country{USA}
}
\email{2396876@students.wits.ac.za}

\author{Ahmed Khalifa}
\affiliation{%
  \institution{Nof1}
  \city{New York}
  \state{NY}
  \country{USA}
}
\email{aak538@nyu.edu}

\author{Lisa Soros}
\affiliation{%
  \institution{Nof1}
  \city{New York}
  \state{NY}
  \country{USA}
}
\email{lisa.soros@gmail.com}

\author{Jay Azhang}
\affiliation{%
  \institution{Nof1}
  \city{New York}
  \state{NY}
  \country{USA}
}
\email{j@thenof1.com}

\author{Julian Togelius}
\affiliation{%
  \institution{Nof1}%
  \city{New York}%
  \state{NY}%
  \country{USA}%
}
\email{julian@togelius.com}

\renewcommand{\shortauthors}{Siper et al.}

\begin{abstract}
Genetic Programming yields interpretable programs, but small syntactic mutations can induce large, unpredictable behavioral shifts, degrading locality and sample efficiency. We frame this as an operator-design problem: learn a continuous program space where latent distance has behavioral meaning, then design mutation operators that exploit this structure without changing the evolutionary optimizer.

We make locality measurable by tracking action-level divergence under controlled latent perturbations, identifying an empirical trust region for behavior-local continuous variation. Using a compact trading-strategy DSL with four semantic components (long/short entry and exit), we learn a matching block-factorized embedding and compare isotropic Gaussian mutation over the full latent space to geometry-compiled mutation that restricts updates to semantically paired entry--exit subspaces and proposes directions using a learned flow-based model trained on logged mutation outcomes.

Under identical $(\mu+\lambda)$ evolution strategies and fixed evaluation budgets across five assets, the learned mutation operator discovers strong strategies using an order of magnitude fewer evaluations and achieves the highest median out-of-sample Sharpe ratio. Although isotropic mutation occasionally attains higher peak performance, geometry-compiled mutation yields faster, more reliable progress, demonstrating that semantically aligned mutation can substantially improve search efficiency without modifying the underlying evolutionary algorithm.
\end{abstract}


\begin{teaserfigure}
    \centering
    \begin{subfigure}[t]{.47\linewidth}
        \includegraphics[height=5cm]{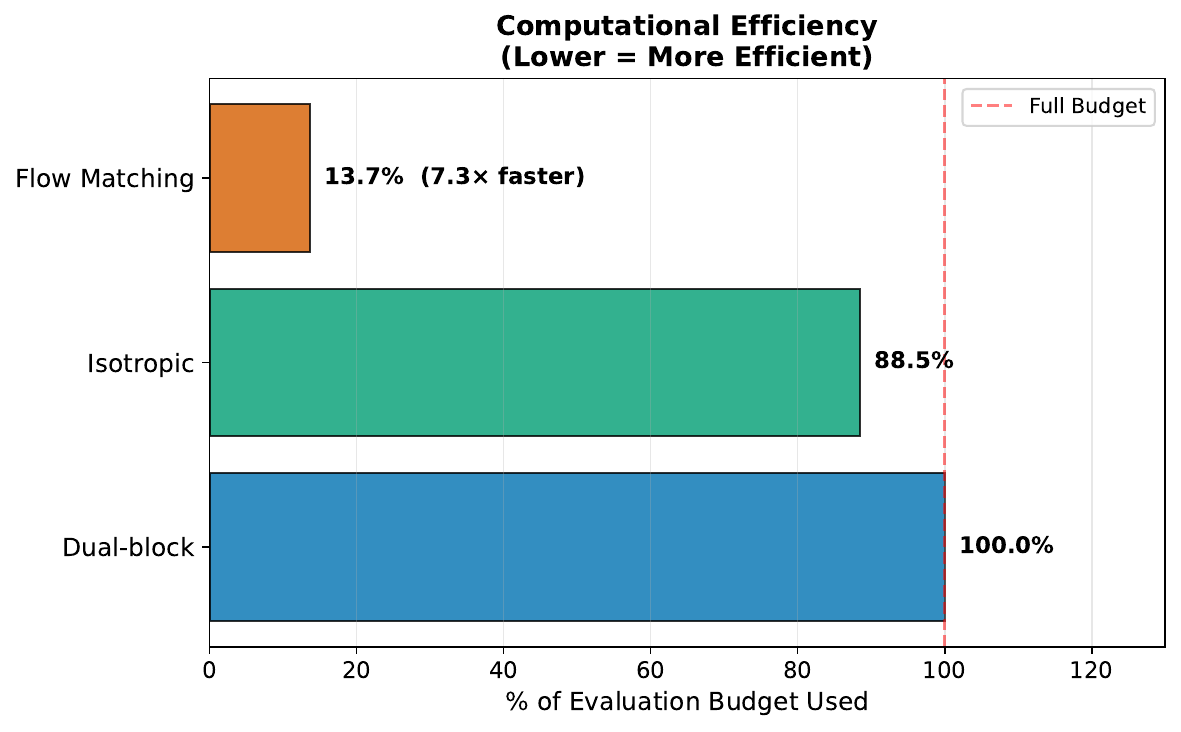}
        \caption{Evaluation budget required to discover strong solutions.}
    \end{subfigure}
    \hfill
    \begin{subfigure}[t]{.47\linewidth}
        \includegraphics[height=5cm]{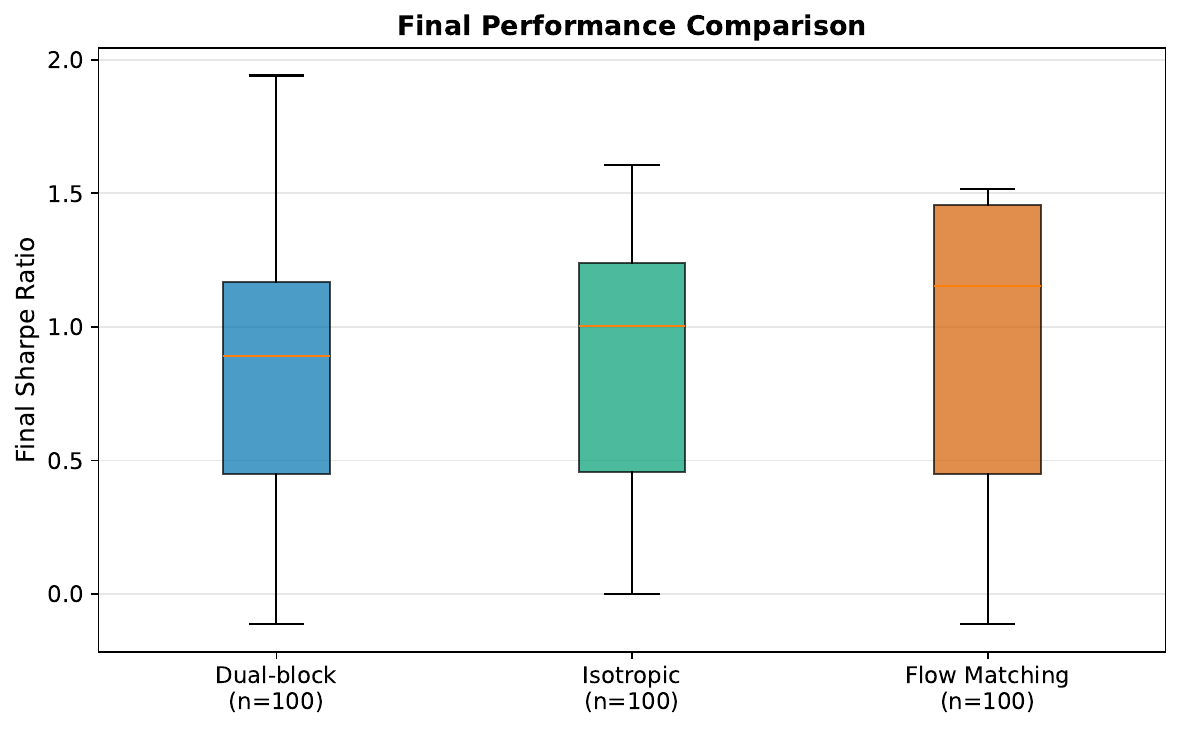}
        \caption{Final out-of-sample Sharpe distributions.}
    \end{subfigure}
\caption{
\textbf{Geometry-aware mutation improves search efficiency and typical performance under a fixed optimizer.}
Both panels compare mutation operators under the same $(\mu+\lambda)$ evolution strategy and identical evaluation budgets.
\textbf{(Left)} shows that geometry-compiled flow-based mutation reaches high-quality strategies using an order of magnitude fewer evaluations than isotropic and dual-block Gaussian baselines.
\textbf{(Right)} reports final out-of-sample Sharpe distributions at the full budget, where the learned operator achieves the highest median performance while remaining competitive in peak outcomes.
}
\label{fig:operator_tradeoff_combined}
\end{teaserfigure}

\maketitle


\section{Introduction}

Symbolic programs are a natural representation for policies and decision rules in program synthesis, control, and algorithmic trading: they are interpretable, compositional, and amenable to grammatical and type-based constraints. A persistent challenge for Genetic Programming (GP), however, is \emph{locality}. Small syntactic edits—such as operator changes or subtree replacements—can induce large, unpredictable behavioral shifts, degrading stability, credit assignment, and sample efficiency. As a result, evolutionary search often expends substantial evaluation budget before discovering useful solutions.

Embedding programs into a continuous search space offers a potential remedy, as it enables incremental variation through small latent perturbations. However, continuous search is only effective when the representation provides a reliable correspondence between \emph{latent distance} and \emph{behavioral change}. When this link is weak, even powerful optimizers struggle: inefficiency arises not from the evolutionary loop itself, but from a mismatch between how variation is applied and how program behavior responds.

We therefore focus on \emph{latent behavioral geometry}: the relationship between movements in latent space and changes in decoded program behavior. Continuous variation is meaningful only within regions where small perturbations produce small, predictable behavioral effects. Outside these regions, decoding failures and unintended cross-talk between program components can render search unstable and inefficient. Understanding and exploiting this geometry is essential for designing effective mutation operators.

Our approach is to make latent behavioral geometry explicit and measurable, then leverage it to design mutation operators under a \emph{fixed} evolutionary algorithm. We first characterize locality using controlled latent perturbations, measuring (i) decode validity, (ii) structural change via normalized AST edit distance, and (iii) behavioral change via action-sequence divergence. Together, these diagnostics identify an empirical trust region in which continuous mutations remain valid and behavior-local.

We then exploit this structure using a compact, typed trading-strategy DSL that decomposes each strategy into four semantic components: long entry (LE), short entry (SE), long exit (LX), and short exit (SX). We learn continuous embeddings with a transformer-based VAE whose latent representation is explicitly factorized along these components. Perturbation and swap tests (Section~\ref{sec:disentanglement}) verify that this factorization provides reliable, low cross-talk control over decoded behavior.

Rather than mutating all latent dimensions simultaneously, we design structured mutation operators that restrict updates to semantically meaningful subspaces. In particular, we exploit the directional structure of trading strategies by pairing long entry with long exit and short entry with short exit. These paired subspaces correspond to coherent trade lifecycles and enable targeted exploration that avoids incoherent offspring which simultaneously alter unrelated signals.

Finally, we ask whether mutation quality can be improved \emph{within} these semantically constrained subspaces. Instead of sampling isotropic noise, we learn a geometry-compiled mutation operator that predicts behavior-improving update directions from logged evolutionary traces. The resulting flow-based mutation model proposes updates in a single forward pass and is applied only through direction-paired masks. Crucially, it serves as a drop-in replacement for standard mutation: the evolutionary loop, selection mechanism, and evaluation budget remain unchanged.

Across multiple assets and fixed evaluation budgets, this learned operator consistently discovers strong strategies using substantially fewer evaluations and achieves higher median out-of-sample performance than unstructured isotropic mutation. While unrestricted mutation occasionally attains higher peak outcomes, geometry-compiled mutation prioritizes efficiency and reliability, yielding faster and more robust evolutionary search.

\paragraph{Contributions.}
This paper makes the following contributions:
\begin{itemize}[leftmargin=*]
\item \textbf{Behavioral geometry diagnostics for program latent spaces.}
We introduce practical measurements of decode validity, structural change, and behavioral divergence to identify trust regions for behavior-local continuous variation.

\item \textbf{A symbolic trading DSL enabling semantically aligned embeddings.}
We design a closed, typed language whose decomposition into entry and exit signals supports interpretable, component-aligned latent representations.

\item \textbf{Structured mutation operators under a fixed optimizer.}
We show that restricting mutation to semantically paired subspaces improves search efficiency and reliability relative to isotropic full-latent mutation, without modifying the evolutionary algorithm.

\item \textbf{A learned geometry-compiled mutation operator.}
We propose a flow-based mutation model trained on logged evolutionary traces that achieves faster discovery and higher median performance under identical $(\mu+\lambda)$ evolution strategies and budgets.
\end{itemize}

Together, these results demonstrate that aligning mutation structure with latent behavioral geometry can substantially improve the efficiency and robustness of continuous program search without changing the underlying evolutionary algorithm.

\section{Related Work}


Our work sits at the intersection of four areas.
These are Genetic Programming and symbolic policy search, continuous and latent-space evolution, disentangled representation learning, and learned variation operators.
Across these areas, many methods rely on some notion of locality.
Our focus is to make locality behavioral, measurable, and directly usable for operator design.

\subsection{Genetic Programming and Symbolic Policy Search}
Genetic Programming (GP) has a long history of producing interpretable programs for program synthesis, control, and decision-making \cite{koza1992programming,o2009riccardo}. A persistent challenge is locality. Small syntactic edits, such as replacing an operator or swapping a subtree, can cause large and hard-to-predict changes in behavior or fitness. This can make evolution unstable and sample-inefficient as programs grow. Many GP systems address this by constraining or structuring the representation. Examples include automatically defined functions \cite{koza1994genetic}, grammar-guided GP \cite{whigham1995grammatically}, and strongly typed GP \cite{montana1995strongly}. These methods improve validity and can make certain edits safer, but the search space remains discrete, and the behavioral meaning of a small syntactic edit is still difficult to quantify.

Other lines of work attack locality more directly through semantics.
Geometric Semantic GP defines operators in semantic space to induce smoother fitness landscapes \cite{moraglio2012geometric}.
Related work also uses semantic information to shape operators and improve search dynamics \cite{pawlak2014semantic}.
These approaches highlight an important point.
Better locality often comes from aligning variation with what the program does, not only with how it is written.

Our approach targets the same locality problem, but through a continuous representation that we evaluate in behavioral terms.
Instead of proposing a new GP optimizer, we hold the optimizer fixed and study how representation geometry and mutation structure affect behavior.
We then use measured latent-to-behavior relationships to design mutation operators that are targeted, behavior-local, and easy to plug into standard evolutionary loops.

\subsection{Continuous and Latent-Space Evolution}

Evolving solutions in continuous spaces is common in neuroevolution and black-box optimization \cite{floreano2008neuroevolution,stanley2019designing}.
Evolution strategies and related methods can be effective when small parameter changes lead to reasonably smooth changes in outcomes. Widely used examples include OpenAI-ES \cite{salimans2017evolution} and CMA-ES \cite{hansen2001completely}. A few recent papers explore embedding programs into continuous spaces and then optimizing those embeddings.
Neural Program Optimization (NPO) trains an autoencoder for programs and uses a continuous optimizer in the latent space, decoding candidates back to programs during search \cite{liskowski2020program}.
Lynch et al.~\cite{lynch2020program} combine grammars with a variational autoencoder to construct a continuous space of programs, then apply evolutionary search in that space.
These papers demonstrate that latent-space search can work for program synthesis and related tasks~\cite{bontrager2018deepmasterprints}.
They also motivate a key practical question.
What does a small latent move actually do to the resulting program?

Our work makes this question explicit and central.
We do not treat the latent space as a black box.
We measure decode validity under perturbation, structural change under perturbation, and behavioral change under perturbation.
This yields an empirical trust region that defines when continuous variation is behaviorally meaningful.
We then hold the optimizer fixed and change only the mutation operator. This isolates the effect of representation geometry and operator design from optimizer tuning.

\subsection{Disentangled Representations and Structured Latents}
Disentangled representation learning aims to separate underlying factors of variation in a way that supports independent control \cite{bengio2013representation}.
In VAEs, common approaches include objectives that encourage factor separation, such as $\beta$-VAE \cite{higgins2017beta} and FactorVAE-style methods \cite{kim2018disentangling}.
The literature also proposes metrics to quantify disentanglement.
Examples include DCI \cite{eastwood2018framework} and MIG \cite{chen2018isolating}, along with critiques that show metrics can disagree and can be brittle in practice \cite{abdi2019preliminary}.

In many optimization settings, disentanglement is treated as a general quality of the embedding.
The hope is that a better representation will make search easier.
In our setting, we use a more operational notion that is directly tied to mutation.
We build an explicitly block-factorized latent aligned to the four semantic strategy parts in our DSL.
We then test whether perturbing one latent block changes only the corresponding decoded program part, including targeted perturbations and swap tests.
This turns disentanglement into a practical resource for operator design rather than a purely descriptive score.
We do not claim full statistical independence.
Our goal is controllable, behavior-local edits that are useful for evolution.

\subsection{Learning Variation Operators and Generative Mutation Models}

Learning better variation is a long-standing theme in evolutionary computation.
Model-based GP and estimation-of-distribution ideas replace hand-designed mutation with learned sampling mechanisms.
A recent example is Denoising Autoencoder Genetic Programming (DAE-GP).
It uses a denoising autoencoder model to generate new candidate programs, and it studies how the corruption process controls exploration and exploitation \cite{wittenberg2023denoising}. Mutation Models is an earlier approach by Khalifa et al.~\cite{khalifa2022mutation} where the mutation function is learned during evolution. This is done by training a small network to learn the mutation function using the evolutionary history while evolution is happening, and then reusing it as the mutation function itself to help direct the search.

We do not learn to generate whole programs directly.
Instead, we learn a conditional distribution over \emph{latent perturbations} inside a measured trust region.
This keeps program validity in the existing decoder and makes the learned model a drop-in replacement for a mutation kernel.
Concretely, we train a conditional flow-based mutation model \cite{ho2020ddpm,song2021ddim} that proposes targeted block-level latent deltas.

\section{Genetic Programming Trading Language (GPTL)}

To study continuous program search in a controlled and reproducible way, we define the \emph{Genetic Programming Trading Language} (GPTL).
GPTL is a small domain-specific language for expressing algorithmic financial trading strategies as symbolic programs.
It serves two purposes.
First, it provides a GP search space with strong safety guarantees.
Second, its structure cleanly separates a strategy into meaningful parts, which later supports a block-factorized continuous representation and block-wise mutation.

\subsection{Language Design Goals}

GPTL was designed with five goals in mind. First, it must be expressive enough to represent a broad set of common technical trading rules, including indicator-based entry and exit logic (a statistical rule-based system that uses price data determine when to buy or sell trades). Second, every generated program must be syntactically and semantically valid. This allows large-scale search without runtime failures caused by malformed programs. Third, program execution must be deterministic and bounded in time and memory. Fourth, programs must have a compact representation that can be embedded by sequence models. Fifth, and most important for this paper, the language must expose a modular strategy structure that makes it possible to treat different parts of a strategy independently during analysis and mutation. To satisfy this final goal, each GPTL strategy is decomposed into four Boolean signal expressions:
\begin{itemize}
    \item \textbf{Long Entry (LE):} if true, buy-to-enter a  long position.
    \item \textbf{Short Entry (SE):} if true, sell-to-enter a short position.
    \item \textbf{Long Exit (LX):} if true, sell-to-close an existing long position.
    \item \textbf{Short Exit (SX):} if true, buy-to-close an existing short position.
\end{itemize}
This decomposition mirrors the lifecycle of a trading position. It also gives us four clearly defined semantic components that can be represented as separate subtrees and, later, as separate latent blocks.

\subsection{Syntax, Grammar, and Type System}
A GPTL signal is a Boolean expression.
Boolean expressions are built from comparisons between numeric expressions, then combined with logical operators. Numeric expressions are formed from three sources. These are price fields, technical indicator calls, and constants. We use a closed, context-free grammar, which means programs can be parsed deterministically, and no out-of-grammar constructs are allowed. GPTL uses a simple static type system with two primitive types. The \texttt{Numeric} type includes price fields, constants, and indicator outputs. The \texttt{Boolean} type includes comparison results and logical expressions. Relational operators map \texttt{Numeric} inputs to \texttt{Boolean}. Logical operators take \texttt{Boolean} inputs.
No implicit type conversions are allowed. All operators have fixed arity and fixed type signatures. All programs are fully parenthesized during serialization to remove ambiguity. During generation and mutation, we enforce bounds on maximum tree depth and minimum structural complexity. These choices guarantee closure under the language rules. Any generated or mutated program is type-correct, syntactically valid, and executable.

\subsection{Program Semantics and Execution Model}
Programs are evaluated over discrete time series containing OHLCV fields (Open, High, Low, Close, and Volume) and precomputed technical indicators. These fields and values give you an indicator of the current state of the financial market at the given time. At each timestep, the four signals (LE, SE, LX, SX) are evaluated independently to Boolean values. Trading follows a deterministic event-driven model.
A long (short) position is opened when the corresponding entry signal is true and no long (short) position is already open. A position is closed when the corresponding exit signal is true. If multiple signals fire at the same timestep, fixed priority rules ensure deterministic behavior. All orders execute at the next-bar open. Given the same market data and parameters, evaluation is fully deterministic. 

\subsection{Abstract Syntax Tree Representation}

Internally, each GPTL strategy is represented as a typed abstract syntax tree (AST).
Leaf nodes correspond to numeric terminals such as price fields and constants.
Internal nodes correspond to indicator calls, relational operators, and logical operators.
The four signal expressions are stored as four disjoint subtrees under dedicated signal roots.
This 
enables modular discrete mutation in program space and later enables a block-structured continuous representation aligned to the same four components.

\subsection{Discrete Mutation Operators}
GPTL includes a set of type-preserving mutation operators defined over the AST. These operators include subtree replacement, operator mutation, terminal mutation, and insertion or deletion (subjected to depth constraints). All operators preserve syntactic validity, type correctness, and grammar closure by construction. Mutations are local, bounded, and well-defined. They provide a  baseline for discrete program-space evolutionary search. They also serve as a reference point for the continuous latent-space mutation operators studied later, which allows us to isolate the effects of representation geometry from the effects of the optimizer.

\section{Learning Continuous Program Embeddings}

To search over programs using continuous mutation, we need a continuous representation of a GPTL strategy. We learn this representation by training a transformer-based variational autoencoder (VAE) that maps a symbolic strategy to a real-valued latent vector and back again.
The model is designed to match the structure of GPTL.
Each strategy is composed of four signals (LE, SE, LX, SX), and our latent representation is explicitly organized into four corresponding latent blocks.
This gives us a representation that is easy to interpret at a high level and is directly usable for block-wise mutation during evolution.

\subsection{Transformer VAE Architecture}

Each GPTL strategy consists of four Boolean signal expressions.
These are long entry (LE), short entry (SE), long exit (LX), and short exit (SX).
We encode and decode each signal separately.
Concretely, each signal expression is serialized into a token sequence and passed through a shared transformer encoder.
The encoder outputs a latent distribution for that signal.
We then concatenate the four signal latents to obtain the full strategy embedding. This process of dividing the latent is called latent factorization.

Let $z_{le}, z_{se}, z_{lx}, z_{sx}$ denote the latent vectors for the four signals.
The full strategy latent is
\[
z = [z_{le}, z_{se}, z_{lx}, z_{sx}] \in \mathbb{R}^{d_{\text{latent}}^{\text{total}}},
\]
where $d_{\text{latent}}^{\text{total}} = 4 \times d_{\text{latent}}^{\text{signal}}$.
This block structure aligns the representation with the four semantic components of the strategy and enables block-wise mutation operators during evolutionary search. Token sequences are embedded into a $d_{\text{model}} = 512$ dimensional space and augmented with sinusoidal positional encodings. The encoder uses four transformer encoder layers.
Each layer contains multi-head self-attention with eight heads and a feedforward network with hidden dimension 1024.
Residual connections, layer normalization, and dropout with probability 0.1 are applied throughout. To produce a single vector per signal, we aggregate the encoder outputs using masked mean pooling over non-padding tokens.
This pooled vector is projected into the parameters of a diagonal Gaussian latent distribution $(\mu, \log \sigma^2)$ for that signal. The decoder mirrors the encoder with four transformer decoder layers. Each signal latent vector is projected to the model dimension and provided to the decoder through cross-attention as a single-token memory. We decode autoregressively with causal masking to generate the token sequence corresponding to the GPTL expression. A final linear projection maps decoder states to vocabulary logits.

\subsection{Tokenization and Serialization}

We use a grammar-aware tokenizer tailored to GPTL.
The vocabulary includes special tokens (PAD, SOS, EOS, UNK), logical and relational operators, price fields, indicator names, punctuation symbols, and discretized numeric tokens.
Numeric values are represented with fixed tokens such as \texttt{<NUM:20>} drawn from a bounded set.
All programs are fully parenthesized during serialization to ensure deterministic parsing and remove ambiguity.

\subsection{Training Procedure}
We generate a training set by randomized GPTL program synthesis.
Each strategy contains four independently generated signal trees represented as ASTs. We use a maximum depth of eight and enforce minimum depth constraints and valid operator arity to avoid degenerate expressions. Each generated strategy is compiled and evaluated on historical market data. Strategies that fail to compile or that produce no trades in any walk-forward fold are discarded. This filtering step removes inactive or ill-formed strategies. It does not apply any selection pressure toward profitability or any specific trading behavior. The final dataset contains 20{,}000 valid strategies. We split it into 80\% training and 20\% validation using a fixed random seed per model replicate.

\paragraph{Objective and optimization.}
We train the VAE with teacher forcing using a reconstruction loss plus a KL regularizer,
\[
\mathcal{L} = \mathcal{L}_{\text{recon}} + \beta \mathcal{L}_{\text{KL}},
\]
where $\mathcal{L}_{\text{recon}}$ is token-level cross-entropy averaged across all four signals, and
\[
\mathcal{L}_{\text{KL}} = -\tfrac{1}{2}\mathbb{E}\bigl[1 + \log \sigma^2 - \mu^2 - \sigma^2\bigr].
\]
We linearly anneal the KL weight $\beta$ from 0 to 0.1 over the first 50 training epochs to mitigate posterior collapse.

We optimize using a weighted decay Adam optimizer (AdamW) with learning rate $10^{-4}$ and weight decay $10^{-5}$. We use cosine learning-rate annealing, gradient clipping at norm 1.0, and automatic mixed precision. All models are trained for 50 epochs with a batch size 128.
We retain the checkpoint with the lowest validation loss.



\section{Latent Quality and Behavioral Geometry Diagnostics}
This section answers two questions. First, does the VAE provide a usable continuous representation of programs, and second, when we take a small step in latent space, what happens to the decoded program? These questions determine whether continuous mutation is meaningful and what mutation scales are safe. We tested the quality across different latent dimension sizes by training models with $d_{\text{latent}}^{\text{total}} \in \{16, 32, 64, 128, 256, 512, 1024, 2048\}$,
where for each latent dimension, we train five independent VAE models using different random initialization seeds.

\subsection{Latent Quality Across Dimensions}
We evaluate VAE quality across the latent-dimension sizes. For each latent size, we train five independent models and report aggregated results. Reconstruction accuracy measures strict round-trip fidelity, counted correctly only if all four signals match exactly after AST canonicalization. Normalized edit distance captures token-level distance between original and reconstruction, averaged across signals. Validity is the fraction of prior samples that decode into executable programs. Uniqueness detects mode collapse; novelty detects memorization. Table~\ref{tab:latent_quality_full} reveals a capacity trade-off: very small latents reconstruct poorly, very large latents decode less robustly, and intermediate sizes provide the best balance. We use intermediate sizes in subsequent analyses.

\subsection{Locality and Behavioral Geometry}
\label{sec:behavioral_geometry}

We measure latent behavioral geometry by asking: if we perturb a latent vector by a controlled amount, how does it change the program's behavior? For each held-out strategy, we encode using the posterior mean and apply an isotropic Gaussian perturbation,
\[
z' = z + \epsilon \eta, \quad \eta \sim \mathcal{N}(0, I),
\]
with $\epsilon \in \{0.01, 0.05, 0.1, 0.5, 1.0, 1.5, 2.5, 3.0, 3.5, 5.0\}$. For each perturbed latent, we measure three quantities: decode success (fraction producing valid programs), structural change (normalized AST edit distance), and behavioral change (action-sequence divergence on shared market data).

\begin{figure}[t]
\centering
\includegraphics[width=0.9\linewidth]{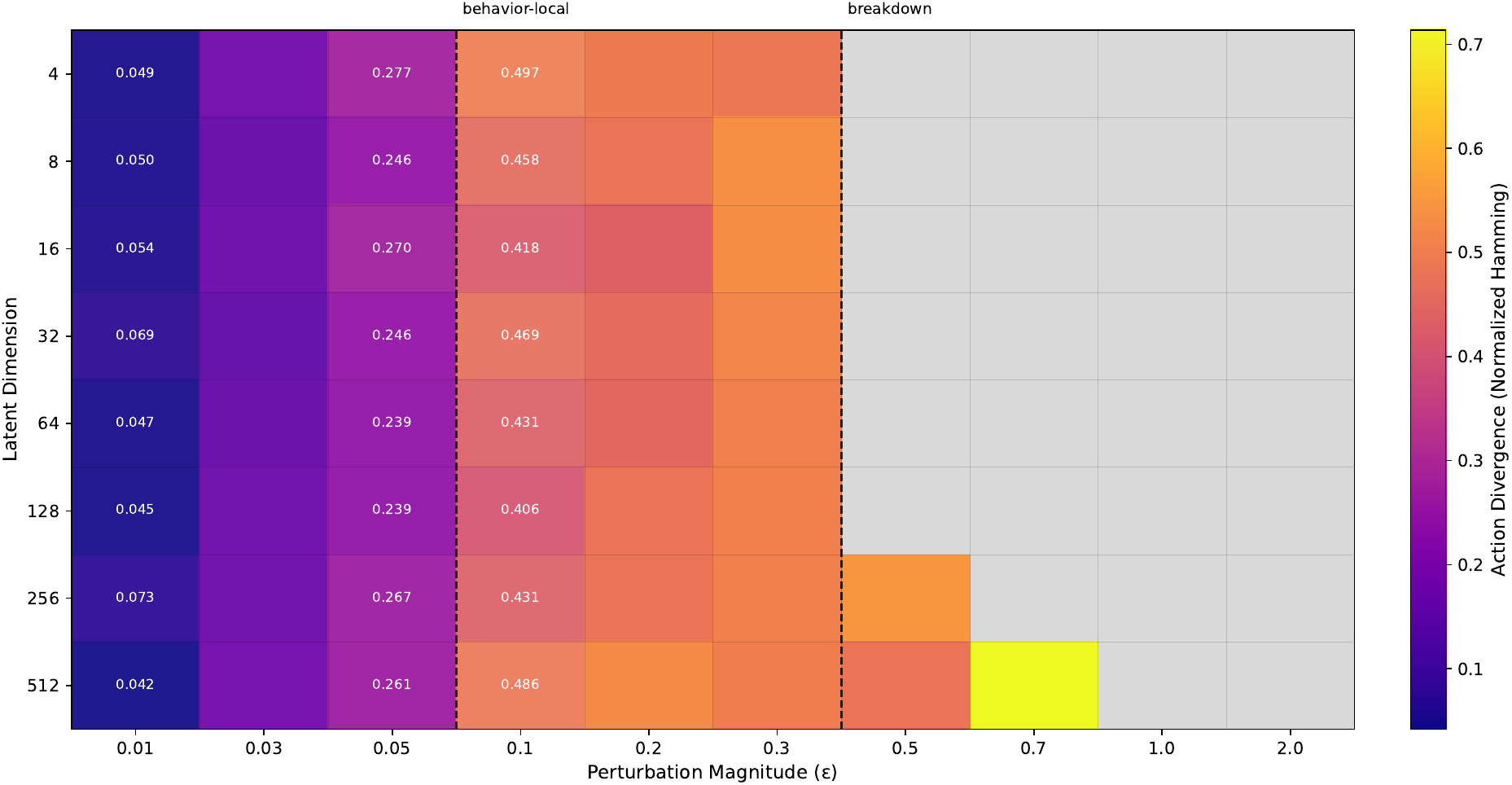}
\caption{Action-sequence divergence between parent and perturbed strategies. Dark colors indicate behavior-local edits. For $\epsilon \leq 0.1$, divergence remains low, defining a trust region. Beyond $\epsilon \geq 0.5$, divergence increases sharply and decoding often fails (gray). This boundary motivates mutation scales in later experiments.}
\label{fig:action_divergence_heatmap}
\end{figure}

Figure~\ref{fig:action_divergence_heatmap} reports behavioral divergence as a function of $\epsilon$ and latent dimensionality. Decode success and behavioral locality remain high for small perturbations ($\epsilon \leq 0.1$), indicating that each strategy has a neighborhood where decoding is stable, and behavior changes predictably. Beyond $\epsilon \geq 0.5$, decode success drops, structural distance grows, and behavioral divergence increases sharply. We treat this transition as an empirical \emph{trust region} for continuous search: inside it, small latent edits yield predictable behavioral refinements; outside, mutations become unreliable. This motivates the geometry-aware operators in ~\ref{sec:gcm}.

\section{Disentanglement and Latent Factorization}
\label{sec:disentanglement}

The mutation operators in this paper rely on a simple idea.
A strategy has four meaningful parts (LE, SE, LX, SX).
Our latent representation is explicitly split into four corresponding blocks.
For block-wise mutation to be reliable, changing one latent block should mainly change the matching strategy part, without unintentionally changing the other three parts.

In this section, we test that this is true in practice.
We treat \emph{latent factorization} as the representation design.
We treat \emph{signal-level disentanglement} as an empirical property of the trained model.
We present two direct tests.
The first asks whether small changes to one latent block stay confined to the intended signal.
The second asks whether we can swap blocks between two strategies and transfer only the corresponding signal.
Together, these tests show that the representation supports predictable, component-level control, which is the key requirement for the mutation operators studied later. These tests both prove disentanglement and zero cross-talk, respectively, and are presented in \ref{sec:signal_disentanglement}

\section{Geometry-Aware Mutation Operators under a Fixed Optimizer}
\label{sec:disentanglement_aware_search}

This section evaluates three mutation operators for continuous program search. All methods use the same evolutionary loop, the same evaluation budget, and the same program embedding model.
The only difference is how new candidate latents are proposed.
This isolates the effect of mutation design from optimizer choice. All flow-based mutations (GCM) are generated using the inference procedure described in Algorithm~\ref{alg:dbd_flow_inference}.

\subsection{Experimental Setup}
\label{sec:exp_setup}
All methods use the same $(\mu+\lambda)$ evolution strategy (ES) in the learned latent space. We fix $\mu=34$ and $\lambda=66$.
Each run evaluates a total of $1320$ offspring, which corresponds to $100$ generations of $\lambda$ offspring. Each candidate latent is decoded into a GPTL program and evaluated by deterministic backtesting. We report out-of-sample Sharpe on held-out test data under a walk-forward protocol explained in the following section. We aggregate results over multiple independent seeds. All mutation operators act in the same 128-dimensional latent space. The isotropic baseline perturbs the full latent vector, while the proposed operator restricts updates to semantically paired entry–exit subspaces corresponding to long-side or short-side trading logic. 

\subsection{Data Splits, Evaluation Protocol, and Trading Assumptions}

All experiments use a fixed five-fold walk-forward evaluation protocol spanning 2008--2025. Each fold consists of approximately 2.5 years of training data, followed by a validation window of approximately 6.5 months and an out-of-sample test window of approximately 6 months. A strict 10-day embargo is enforced between consecutive splits to prevent lookahead bias. All reported results are based exclusively on test folds, and the same splits are used across all assets, mutation operators, and random seeds (see Table~\ref{tab:folds}). Strategies are evaluated using a deterministic backtesting engine with an initial equity of \$10{,}000. Only one position may be held at a time, and position scaling is not allowed. Orders execute at the next-bar open with a transaction slippage cost of 0.1\% and exchange fees of 0.05\% per trade. Experiments are conducted independently on five liquid futures contracts: S\&P~500 (ES), Natural Gas (NG), Crude Oil (CL), Silver (SI), and Euro FX (E6). No model selection, hyperparameter tuning, or operator design choices use information from the test folds.

\subsection{Isotropic Gaussian Mutation}
\label{sec:baseline_latent_search}
The isotropic baseline applies Gaussian mutation uniformly across the entire latent space. Given a parent latent $\mathbf{z}\in\mathbb{R}^{128}$, offspring are generated as
\[
\mathbf{z}' = \mathbf{z} + \sigma\,\boldsymbol{\epsilon}, \qquad \boldsymbol{\epsilon}\sim\mathcal{N}(\mathbf{0},\mathbf{I}_{128}),
\]
so that all 128 latent dimensions are perturbed simultaneously without regard to semantic signal boundaries. This baseline reflects standard practice in latent-space evolutionary search and serves as an unstructured exploration reference.

\subsection{Geometry-Compiled Mutation via Dual-Block Directional Flow (GCM)}
\label{sec:gcm}

We use the term \emph{geometry-compiled mutation} (GCM) to refer to the general approach of learning behavior-aware mutation proposals from logged evolutionary traces; in this work, GCM is instantiated using a flow-matching model. The isotropic Gaussian baseline perturbs all 128 latent dimensions simultaneously, without regard to trading semantics.
We now introduce a learned mutation operator that restricts updates to semantically meaningful subspaces while improving proposal quality within those subspaces.
We refer to this operator as \emph{geometry-compiled mutation} (GCM).

\paragraph{Dual-block directional mutation.}
Trading strategies exhibit directional structure: long positions require coordinated entry and exit rules, as do short positions.
Accordingly, we define two direction-paired mutation subspaces.
Long Entry (LE) and Long Exit (LX) form the \emph{long-side} pair, while Short Entry (SE) and Short Exit (SX) form the \emph{short-side} pair.
Let the latent be ordered as
\[
\mathbf{z} = [\mathbf{z}^{(\mathrm{LE})}, \mathbf{z}^{(\mathrm{SE})}, \mathbf{z}^{(\mathrm{LX})}, \mathbf{z}^{(\mathrm{SX})}] \in \mathbb{R}^{128},
\]
with each block in $\mathbb{R}^{32}$.
We define two non-contiguous binary masks,
\[
\mathbf{m}_{\text{long}} = [\mathbf{1}_{32}, \mathbf{0}_{32}, \mathbf{1}_{32}, \mathbf{0}_{32}], \qquad
\mathbf{m}_{\text{short}} = [\mathbf{0}_{32}, \mathbf{1}_{32}, \mathbf{0}_{32}, \mathbf{1}_{32}],
\]
which activate the long-side and short-side subspaces respectively.
At each generation, the mutation direction alternates deterministically between these two masks.

\paragraph{Learned mutation proposal.}
Rather than sampling isotropic noise within the active subspace, GCM uses a learned flow-matching model to predict a behavior-improving update direction.
The model predicts a velocity field
\[
\mathbf{v}_\theta(\mathbf{z}, \boldsymbol{\phi}) \in \mathbb{R}^{128},
\]
conditioned on the full parent latent $\mathbf{z}$ and an 8-dimensional behavioral embedding $\boldsymbol{\phi}$ computed from the parent strategy’s execution trace.
Unlike diffusion-based approaches, this formulation requires no iterative denoising or timestep conditioning: the update direction is predicted in a single forward pass.

At search time, the predicted velocity is applied only through the active direction-paired mask,
\[
\mathbf{z}' = \mathbf{z} + \alpha \, \mathbf{m}_{d(g)} \odot \mathbf{v}_\theta(\mathbf{z}, \boldsymbol{\phi}) + \boldsymbol{\eta},
\qquad
\boldsymbol{\eta} \sim \mathbf{m}_{d(g)} \odot \mathcal{N}(\mathbf{0}, \sigma^2 \mathbf{I}),
\]
where $d(g) \in \{\text{long}, \text{short}\}$ denotes the active direction at generation $g$.
Dual-block mutation updates 64 of the 128 latent dimensions per generation, focusing exploration on a coherent trade lifecycle rather than perturbing all signals simultaneously.

\paragraph{Training data and usage.}
The flow model is trained offline using logged mutation traces collected under the same $(\mu+\lambda)$ evolution strategy used in our main experiments, but with isotropic Gaussian mutation as the proposal mechanism.
We collect traces across five assets and five independent runs per asset.
Each run evaluates $\lambda=66$ offspring per generation for $100$ generations, yielding $6600$ mutation attempts per run.
Each mutation record contains the parent latent $\mathbf{z}$, the child latent $\mathbf{z}'$, the associated behavioral embedding $\boldsymbol{\phi}$, validity flags, and parent and child fitness values.
During evolution, GCM is used solely as a drop-in replacement for the mutation kernel; the optimizer, evaluation budget, and selection procedure remain unchanged. Algorithm~\ref{alg:dbd_flow_inference} summarizes the inference-time mutation procedure used by the geometry-compiled flow operator within the $(\mu+\lambda)$ evolution strategy.

\begin{algorithm}[t]
\caption{DBD-Flow mutation (inference-time) within $(\mu+\lambda)$-ES}
\label{alg:dbd_flow_inference}
\begin{algorithmic}[1]
\REQUIRE Flow model $F_\theta(\mathbf{z},\boldsymbol{\phi}) \rightarrow \boldsymbol{\delta}_{\text{full}}$ (predicts a 128-dim improvement delta)
\REQUIRE Parent latent $\mathbf{z}\in\mathbb{R}^{128}$, behavioral embedding $\boldsymbol{\phi}\in\mathbb{R}^{8}$, generation index $g$
\REQUIRE Delta scale $\alpha$ (default $1.0$), input noise $\sigma_{\text{in}}$ (default $0.0$), output noise $\sigma_{\text{out}}$ (default $0.0$)
\REQUIRE Masks $\mathbf{m}_{\text{long}}=[\mathbf{1}_{32},\mathbf{0}_{32},\mathbf{1}_{32},\mathbf{0}_{32}]$ and $\mathbf{m}_{\text{short}}=[\mathbf{0}_{32},\mathbf{1}_{32},\mathbf{0}_{32},\mathbf{1}_{32}]$
\ENSURE Mutated latent $\mathbf{z}'\in\mathbb{R}^{128}$

\STATE \textbf{Select direction-paired mask:}
\IF{$g \bmod 2 = 0$}
    \STATE $\mathbf{m} \leftarrow \mathbf{m}_{\text{long}}$ \COMMENT{Mutate (LE, LX)}
\ELSE
    \STATE $\mathbf{m} \leftarrow \mathbf{m}_{\text{short}}$ \COMMENT{Mutate (SE, SX)}
\ENDIF

\STATE \textbf{Optional input exploration:}
\STATE Sample $\boldsymbol{\epsilon}\sim \mathcal{N}(\mathbf{0},\mathbf{I}_{128})$
\STATE $\tilde{\mathbf{z}} \leftarrow \mathbf{z} + \sigma_{\text{in}}\boldsymbol{\epsilon}$

\STATE \textbf{Predict full delta (one forward pass):}
\STATE $\boldsymbol{\delta}_{\text{full}} \leftarrow F_\theta(\tilde{\mathbf{z}}, \boldsymbol{\phi})$

\STATE \textbf{Mask to the active direction-paired subspace:}
\STATE $\boldsymbol{\delta}_{\text{masked}} \leftarrow \mathbf{m}\odot \boldsymbol{\delta}_{\text{full}}$

\STATE \textbf{Optional output noise (active blocks only):}
\STATE Sample $\boldsymbol{\epsilon}'\sim \mathcal{N}(\mathbf{0},\mathbf{I}_{128})$
\STATE $\boldsymbol{\delta} \leftarrow \alpha\,\boldsymbol{\delta}_{\text{masked}} + \sigma_{\text{out}}(\mathbf{m}\odot \boldsymbol{\epsilon}')$

\STATE \textbf{Apply mutation:}
\STATE $\mathbf{z}' \leftarrow \mathbf{z} + \boldsymbol{\delta}$

\RETURN $\mathbf{z}'$
\end{algorithmic}
\end{algorithm}

\subsection{Controlled Comparison}
\label{sec:controlled_ablation}

We compare mutation operators under identical $(\mu+\lambda)$ evolution strategy settings, evaluation budgets, and deterministic seeds to isolate the effect of mutation design.
Specifically, we evaluate:
\begin{enumerate}[leftmargin=*]
\item \textbf{Isotropic Gaussian mutation}, which perturbs all 128 latent dimensions simultaneously without regard to semantic structure;
\item \textbf{Dual-block Gaussian mutation}, which restricts updates to semantically paired entry--exit subspaces (long-side or short-side) using isotropic noise;
\item \textbf{Geometry-compiled mutation (GCM)}, which uses a learned flow-based model to propose update directions within the same dual-block subspaces.
\end{enumerate}

All methods are evaluated across five assets using the same evaluation budget.
Results are reported exclusively on out-of-sample test folds.
Table~\ref{tab:final_results} summarizes median and maximum Sharpe ratios along with evaluation budget usage.

\begin{table}[t]
\centering
\caption{Median and maximum out-of-sample Sharpe ratios and evaluation budget usage across all assets. Lower values of \textit{Pct. Budget Used} indicate faster discovery of strong solutions.}
\label{tab:final_results}
\begin{tabular}{lccc}
\toprule
Method & Median Sharpe & Max Sharpe & Pct. Budget Used \\
\midrule
Flow (GCM) & \textbf{1.152} & 1.518 & \textbf{13.7} \\
Isotropic  & 1.005 & 1.607 & 88.5 \\
Dual-block & 0.890 & \textbf{1.941} & 100.0 \\
\bottomrule
\end{tabular}
\end{table}

\subsection{Interpretation}

The results reveal a clear trade-off between unstructured exploration and semantically aligned mutation.
Isotropic Gaussian mutation occasionally attains high peak Sharpe values, but typically requires most of the evaluation budget and exhibits high variance across runs.
In contrast, restricting mutation to semantically meaningful entry--exit subspaces substantially improves search efficiency and reliability.

The learned geometry-compiled mutation operator achieves the highest median out-of-sample Sharpe while using an order of magnitude less evaluation budget than both isotropic and dual-block Gaussian baselines. This indicates that learning behavior-aware mutation directions prioritizes consistent improvement over rare, high-variance outcomes.
Although GCM does not always achieve the highest absolute Sharpe, it discovers strong strategies much earlier and with greater consistency across assets and seeds.

Taken together, these results show that the primary benefit of geometry-aware mutation lies in faster and more reliable discovery rather than extreme peak optimization.

\section{Discussion}

This work examines continuous program search through the lens of \emph{latent behavioral geometry}. Rather than assuming that continuous embeddings yield smooth or well-behaved search landscapes, we explicitly measure how latent perturbations translate into decoded program behavior. These measurements identify trust regions in which continuous mutation remains valid and behavior-local, providing a principled basis for mutation operator design.

Our results show that performance differences arise primarily from how mutation operators interact with this geometry, not from changes to the evolutionary optimizer. Unstructured isotropic mutation perturbs all signals simultaneously, often producing incoherent edits and requiring large evaluation budgets. Restricting mutation to semantically aligned entry--exit subspaces substantially improves efficiency by focusing exploration on coherent trade lifecycles. Learning mutation directions within these subspaces further accelerates discovery by biasing search toward behaviorally meaningful changes.

Importantly, the learned operator replaces only the mutation kernel. Selection, evaluation, and population dynamics are unchanged. The resulting gains therefore reflect more effective use of limited evaluation budgets rather than increased optimizer complexity or computational effort. While our experiments are conducted in a trading domain, the underlying principles are more general. Many GP systems rely on mutation operators whose behavioral effects are difficult to predict. Our results suggest that measuring latent behavioral geometry and aligning mutation structure with known semantic decompositions can significantly improve the efficiency and reliability of evolutionary search.

There are limitations to this approach. Our DSL admits a clear semantic decomposition into entry and exit signals; other domains may require different representations. Continuous mutation remains effective only within trust regions that preserve validity, limiting exploration radius. Extending these ideas to domains without obvious semantic partitions remains an important direction for future work.

\section{Conclusion}
Continuous program embeddings are only useful when small latent changes correspond to small, interpretable behavioral effects.
Rather than assuming this property, we measured it directly using controlled perturbations that track decode validity, structural change, and behavioral divergence. These diagnostics identify trust regions where continuous mutation is reliable.

We used this information to design semantically aligned mutation operators under a fixed evolutionary algorithm.
Restricting mutation to paired entry-exit subspaces improves efficiency and reliability relative to isotropic full-latent mutation. Learning geometry-compiled mutation directions within these subspaces further accelerates discovery, yielding higher median out-of-sample performance while using substantially less evaluation budget. The central takeaway is that aligning mutation structure with latent behavioral geometry can trade rare peak outcomes for faster, more robust evolutionary search---without modifying the underlying evolutionary algorithm.

\bibliographystyle{acmart}
\bibliography{references}

@book{koza1992programming,
  title        = {Genetic Programming: On the Programming of Computers by Means of Natural Selection},
  author       = {Koza, John R.},
  year         = {1992},
  publisher    = {MIT Press},
  address      = {Cambridge, MA}
}

@inproceedings{bontrager2018deepmasterprints,
  title={Deepmasterprints: Generating masterprints for dictionary attacks via latent variable evolution},
  author={Bontrager, Philip and Roy, Aditi and Togelius, Julian and Memon, Nasir and Ross, Arun},
  booktitle={2018 IEEE 9th International Conference on Biometrics Theory, Applications and Systems (BTAS)},
  pages={1--9},
  year={2018},
  organization={IEEE}
}

@book{koza1994genetic,
  title        = {Genetic Programming II: Automatic Discovery of Reusable Programs},
  author       = {Koza, John R.},
  year         = {1994},
  publisher    = {MIT Press},
  address      = {Cambridge, MA}
}

@inproceedings{khalifa2022mutation,
  title={Mutation models: Learning to generate levels by imitating evolution},
  author={Khalifa, Ahmed and Togelius, Julian and Green, Michael Cerny},
  booktitle={Proceedings of the 17th International Conference on the Foundations of Digital Games},
  pages={1--9},
  year={2022}
}

@inproceedings{whigham1995grammatically,
  title        = {Grammatically-based Genetic Programming},
  author       = {Whigham, Peter A.},
  booktitle    = {Proceedings of the Workshop on Genetic Programming: From Theory to Real-World Applications},
  year         = {1995}
}

@article{abdi2019preliminary,
  title={A preliminary study of disentanglement with insights on the inadequacy of metrics},
  author={Abdi, Amir H and Abolmaesumi, Purang and Fels, Sidney},
  journal={arXiv preprint arXiv:1911.11791},
  year={2019}
}

@article{montana1995strongly,
  title        = {Strongly Typed Genetic Programming},
  author       = {Montana, David J.},
  journal      = {Evolutionary Computation},
  volume       = {3},
  number       = {2},
  pages        = {199--230},
  year         = {1995}
}

@inproceedings{moraglio2012geometric,
  title        = {Geometric Semantic Genetic Programming},
  author       = {Moraglio, Alberto and Krawiec, Krzysztof and Johnson, Colin G.},
  booktitle    = {Parallel Problem Solving from Nature (PPSN)},
  pages        = {21--31},
  year         = {2012},
  organization = {Springer}
}

@article{pawlak2014semantic,
  title        = {Semantic Backpropagation for Designing Search Operators in Genetic Programming},
  author       = {Pawlak, Tomasz P. and Wieloch, Bartosz and Krawiec, Krzysztof},
  journal      = {IEEE Transactions on Evolutionary Computation},
  volume       = {19},
  number       = {3},
  pages        = {326--340},
  year         = {2014},
  publisher    = {IEEE}
}

@article{floreano2008neuroevolution,
  title        = {Neuroevolution: From Architectures to Learning},
  author       = {Floreano, Dario and D{\"u}rr, Peter and Mattiussi, Claudio},
  journal      = {Evolutionary Intelligence},
  volume       = {1},
  number       = {1},
  pages        = {47--62},
  year         = {2008},
  publisher    = {Springer}
}

@article{stanley2019designing,
  title        = {Designing Neural Networks through Neuroevolution},
  author       = {Stanley, Kenneth O. and Clune, Jeff and Lehman, Joel and Miikkulainen, Risto},
  journal      = {Nature Machine Intelligence},
  volume       = {1},
  number       = {1},
  pages        = {24--35},
  year         = {2019},
  publisher    = {Nature Publishing Group}
}

@article{hansen2001completely,
  title        = {Completely Derandomized Self-Adaptation in Evolution Strategies},
  author       = {Hansen, Nikolaus and Ostermeier, Andreas},
  journal      = {Evolutionary Computation},
  volume       = {9},
  number       = {2},
  pages        = {159--195},
  year         = {2001},
  publisher    = {MIT Press}
}

@article{salimans2017evolution,
  title        = {Evolution Strategies as a Scalable Alternative to Reinforcement Learning},
  author       = {Salimans, Tim and Ho, Jonathan and Chen, Xi and Sidor, Szymon and Sutskever, Ilya},
  journal      = {arXiv preprint arXiv:1703.03864},
  year         = {2017}
}

@inproceedings{liskowski2020program,
  title        = {Program Synthesis as Latent Continuous Optimization: Evolutionary Search in Neural Embeddings},
  author       = {Liskowski, Pawe{\l} and Krawiec, Krzysztof and Toklu, Nihat Engin and Swan, Jerry},
  booktitle    = {Proceedings of the 2020 Genetic and Evolutionary Computation Conference},
  pages        = {359--367},
  year         = {2020},
  organization = {ACM},
  doi          = {10.1145/3377930.3390213}
}

@inproceedings{lynch2020program,
  title        = {Program Synthesis in a Continuous Space Using Grammars and Variational Autoencoders},
  author       = {Lynch, David and McDermott, James and O'Neill, Michael},
  booktitle    = {Parallel Problem Solving from Nature (PPSN XVI)},
  series       = {Lecture Notes in Computer Science},
  volume       = {12270},
  pages        = {33--47},
  year         = {2020},
  publisher    = {Springer},
  doi          = {10.1007/978-3-030-58115-2_3}
}

@article{bengio2013representation,
  title        = {Representation Learning: A Review and New Perspectives},
  author       = {Bengio, Yoshua and Courville, Aaron and Vincent, Pascal},
  journal      = {IEEE Transactions on Pattern Analysis and Machine Intelligence},
  volume       = {35},
  number       = {8},
  pages        = {1798--1828},
  year         = {2013},
  publisher    = {IEEE}
}

@inproceedings{higgins2017beta,
  title        = {beta-VAE: Learning Basic Visual Concepts with a Constrained Variational Framework},
  author       = {Higgins, Irina and Matthey, Loic and Pal, Arka and Burgess, Christopher and Glorot, Xavier and Botvinick, Matthew and Mohamed, Shakir and Lerchner, Alexander},
  booktitle    = {International Conference on Learning Representations},
  year         = {2017}
}

@inproceedings{kim2018disentangling,
  title        = {Disentangling by Factorising},
  author       = {Kim, Hyunjik and Mnih, Andriy},
  booktitle    = {International Conference on Machine Learning},
  pages        = {2649--2658},
  year         = {2018},
  organization = {PMLR}
}

@article{chen2018isolating,
  title        = {Isolating Sources of Disentanglement in Variational Autoencoders},
  author       = {Chen, Ricky T. Q. and Li, Xuechen and Grosse, Roger B. and Duvenaud, David K.},
  journal      = {Advances in Neural Information Processing Systems},
  volume       = {31},
  year         = {2018}
}

@inproceedings{eastwood2018framework,
  title        = {A Framework for the Quantitative Evaluation of Disentangled Representations},
  author       = {Eastwood, Cian and Williams, Christopher K. I.},
  booktitle    = {International Conference on Learning Representations},
  year         = {2018}
}

@article{wittenberg2023denoising,
  title        = {Denoising Autoencoder Genetic Programming: Strategies to Control Exploration and Exploitation in Search},
  author       = {Wittenberg, David and Rothlauf, Franz and Gagn{\'e}, Christian},
  journal      = {Genetic Programming and Evolvable Machines},
  volume       = {24},
  number       = {2},
  pages        = {17},
  year         = {2023},
  publisher    = {Springer}
}

@inproceedings{ho2020ddpm,
  title        = {Denoising Diffusion Probabilistic Models},
  author       = {Ho, Jonathan and Jain, Ajay and Abbeel, Pieter},
  booktitle    = {Advances in Neural Information Processing Systems},
  volume       = {33},
  year         = {2020}
}

@inproceedings{song2021ddim,
  title        = {Denoising Diffusion Implicit Models},
  author       = {Song, Jiaming and Meng, Chenlin and Ermon, Stefano},
  booktitle    = {International Conference on Learning Representations},
  year         = {2021}
}

@misc{o2009riccardo,
  title={Riccardo Poli, William B. Langdon, Nicholas F. McPhee: A Field Guide to Genetic Programming: Lulu. com, 2008, 250 pp, ISBN 978-1-4092-0073-4},
  author={O’Neill, Michael},
  year={2009},
  publisher={Springer}
}

\appendix

\section{Appendix}

The appendices provide full technical details necessary for reproducibility and to support the empirical claims made in the main paper. All experiments can be reproduced using the specifications, datasets, and hyperparameters described below.

\section{Full GPTL Grammar and Mutation Specifications}
\label{app:gptl}

\subsection{Complete Grammar}

GPTL programs are composed of four Boolean signal expressions (LE, SE, LX, SX), each defined by the following context-free grammar:

\begin{verbatim}
signal          ::= expr

expr            ::= expr '&' expr
                  | expr '|' expr
                  | '~' expr
                  | comparison

comparison      ::= numeric_expr relop numeric_expr
relop           ::= '>' | '<' | '>=' | '<=' | '=='

numeric_expr    ::= indicator_call | field | constant
indicator_call  ::= IND '(' field ',' INT ')'

field           ::= open | high | low | close | volume
constant        ::= FLOAT
\end{verbatim}

All expressions are fully parenthesized during serialization. Operator precedence is fixed as NOT $>$ AND $>$ OR. The grammar is closed and guarantees syntactic validity.

\subsection{Static Typing Rules}

GPTL enforces a strict static type system with two primitive types:
\begin{itemize}
\item \texttt{Numeric}: price fields, constants, indicator outputs
\item \texttt{Boolean}: comparison results and logical expressions
\end{itemize}

Typing rules prohibit implicit coercions. All operators have fixed arity and type signatures, ensuring closure under mutation.

\subsection{Mutation Operators}

All discrete mutation operators are type-preserving and grammar-safe:
\begin{itemize}
\item \textbf{Subtree replacement:} Replace a randomly selected subtree with a newly sampled compatible subtree.
\item \textbf{Operator mutation:} Replace a logical or comparison operator with another operator of the same arity and type.
\item \textbf{Terminal mutation:} Resample numeric constants or indicator parameters within predefined bounds.
\item \textbf{Insertion/deletion:} Insert or remove subtrees subject to maximum depth constraints.
\end{itemize}

All mutations preserve syntactic validity, type correctness, and bounded execution.


\subsection{Program Dataset Generation}

The GPTL training dataset is generated via randomized program synthesis. Each strategy consists of four independently generated signal trees with:
\begin{itemize}
\item maximum AST depth: 8
\item minimum depth: 2
\item bounded numeric constants and indicator periods
\end{itemize}

Programs are compiled and evaluated on historical data. Strategies that fail to compile or produce zero trades in any walk-forward fold are discarded. No selection pressure toward profitability is applied.



\section{Complete Latent Quality Tables}
\label{app:latent_quality}

Table~\ref{tab:latent_quality_full} reports the full latent quality metrics across all evaluated latent dimensions, including reconstruction accuracy, normalized edit distance, consistency, validity, uniqueness, and novelty. Results are aggregated across five independently trained VAE models per dimension.

\begin{table}[h]
\centering
\caption{Complete latent quality metrics across latent dimensionalities.}
\label{tab:latent_quality_full}
\resizebox{\columnwidth}{!}{%
\begin{tabular}{lcccccc}
\toprule
Latent Dim. & Recon. Acc. & Norm. Edit Dist. & Consistency & Validity & Uniqueness & Novelty \\
\midrule
16   & 0.006 $\pm$ 0.003 & 0.746 $\pm$ 0.022 & 1.000 $\pm$ 0.000 & 0.896 $\pm$ 0.021 & 0.991 $\pm$ 0.006 & 0.952 $\pm$ 0.010 \\
32   & 0.057 $\pm$ 0.007 & 0.596 $\pm$ 0.017 & 1.000 $\pm$ 0.000 & 0.954 $\pm$ 0.019 & 0.999 $\pm$ 0.001 & 0.965 $\pm$ 0.009 \\
64   & 0.242 $\pm$ 0.016 & 0.385 $\pm$ 0.008 & 1.000 $\pm$ 0.000 & 0.975 $\pm$ 0.014 & 1.000 $\pm$ 0.000 & 0.977 $\pm$ 0.005 \\
128  & 0.414 $\pm$ 0.012 & 0.289 $\pm$ 0.007 & 1.000 $\pm$ 0.000 & 0.978 $\pm$ 0.007 & 1.000 $\pm$ 0.000 & 0.974 $\pm$ 0.005 \\
256  & 0.446 $\pm$ 0.017 & 0.278 $\pm$ 0.009 & 1.000 $\pm$ 0.000 & 0.942 $\pm$ 0.025 & 1.000 $\pm$ 0.000 & 0.983 $\pm$ 0.008 \\
512  & 0.496 $\pm$ 0.008 & 0.280 $\pm$ 0.003 & 1.000 $\pm$ 0.000 & 0.926 $\pm$ 0.021 & 1.000 $\pm$ 0.000 & 0.991 $\pm$ 0.007 \\
1024 & 0.504 $\pm$ 0.024 & 0.271 $\pm$ 0.006 & 1.000 $\pm$ 0.000 & 0.909 $\pm$ 0.040 & 1.000 $\pm$ 0.000 & 0.996 $\pm$ 0.004 \\
2048 & 0.519 $\pm$ 0.010 & 0.270 $\pm$ 0.003 & 1.000 $\pm$ 0.000 & 0.893 $\pm$ 0.029 & 1.000 $\pm$ 0.000 & 1.000 $\pm$ 0.000 \\
\bottomrule
\end{tabular}
}%
\end{table}

\section{Additional Disentanglement Visualizations}
\label{app:disentanglement}

This appendix includes additional heatmaps and visualizations supporting Section~6, including:
\begin{itemize}
\item Signal disentanglement heatmaps across multiple latent dimensions
\item Swap test results for all signal pairs
\item Sensitivity of disentanglement metrics to perturbation scale
\end{itemize}

All visualizations are computed using the same evaluation protocols described in the main text.




\subsection{Behavioral Embedding $\Phi$ and Trust-Region Binning}
\label{app:phi_embedding}

GCM conditions on a compact, interpretable behavioral embedding $\Phi(\cdot)\in\mathbb{R}^8$ computed from a strategy's execution trace under deterministic backtesting.
Let $T$ denote the number of evaluated bars in the backtest window.
Let $\text{pos}_t \in \{-1,0,1\}$ denote the position at time $t$, where $1$ indicates long, $-1$ indicates short, and $0$ indicates flat.

\paragraph{Market regime.}
We define a simple market regime indicator using a 100-bar moving average of the close price.
Let $c_t$ be the close at time $t$ and let $\text{MA}_{100}(c)_t$ be the 100-bar moving average.
We define
\[
r_t = \mathbb{I}\left[c_t > \text{MA}_{100}(c)_t\right],
\]
where $r_t=1$ indicates an up regime and $r_t=0$ indicates a down regime.
We compute all regime-based features over timesteps where $\text{MA}_{100}$ is defined.

\paragraph{Regime exposure features.}
Define indicators $\ell_t = \mathbb{I}[\text{pos}_t>0]$ and $s_t = \mathbb{I}[\text{pos}_t<0]$.
The first four components of $\Phi$ measure the fraction of time spent in long or short positions during up or down regimes
\[
\phi_1 = \frac{1}{T}\sum_{t=1}^{T} \ell_t (1-r_t),
\quad
\phi_2 = \frac{1}{T}\sum_{t=1}^{T} s_t (1-r_t),
\]
\[
\phi_3 = \frac{1}{T}\sum_{t=1}^{T} \ell_t r_t,
\quad
\phi_4 = \frac{1}{T}\sum_{t=1}^{T} s_t r_t.
\]

\paragraph{Event statistics.}
We define an entry event as a transition from flat to non-flat.
We define an exit event as a transition from non-flat to flat.
Let $N_{\text{entry}}$ be the number of entry events and $N_{\text{exit}}$ be the number of exit events.
We define normalized entry and exit rates
\[
\phi_5 = \frac{N_{\text{entry}}}{T},
\quad
\phi_6 = \frac{N_{\text{exit}}}{T}.
\]
We also compute hold durations in bars for each completed trade.
Let $\{h_j\}_{j=1}^{J}$ be the set of hold durations for $J$ completed trades.
We define normalized hold statistics
\[
\phi_7 = \frac{\text{mean}(\{h_j\})}{T},
\quad
\phi_8 = \frac{\text{std}(\{h_j\})}{T}.
\]

\paragraph{Behavioral step size, $\rho$ bins, and trust region.}
Given a parent strategy and a decoded child strategy, we compute
\[
\phi_{L2} = \left\lVert \Phi(\text{child}) - \Phi(\text{parent}) \right\rVert_2.
\]
We discretize $\phi_{L2}$ into a four-level bin $\rho$ using fixed thresholds
\[
\rho =
\begin{cases}
\texttt{tiny} & \phi_{L2} \in [0,0.05)\\
\texttt{small} & \phi_{L2} \in [0.05,0.15)\\
\texttt{medium} & \phi_{L2} \in [0.15,0.35)\\
\texttt{large} & \phi_{L2} \in [0.35,\infty).
\end{cases}
\]
We define the behavioral trust region as $\phi_{L2} \le 0.35$.
GCM is trained on mutations within this trust region so that the learned operator focuses on behavior-local edits.
At inference time we typically request $\rho=\texttt{small}$ to remain within the empirically stable locality regime.


\subsection{Signal Disentanglement Test}
\label{sec:signal_disentanglement}
\begin{figure}[t]
\centering
\includegraphics[width=0.65\linewidth]{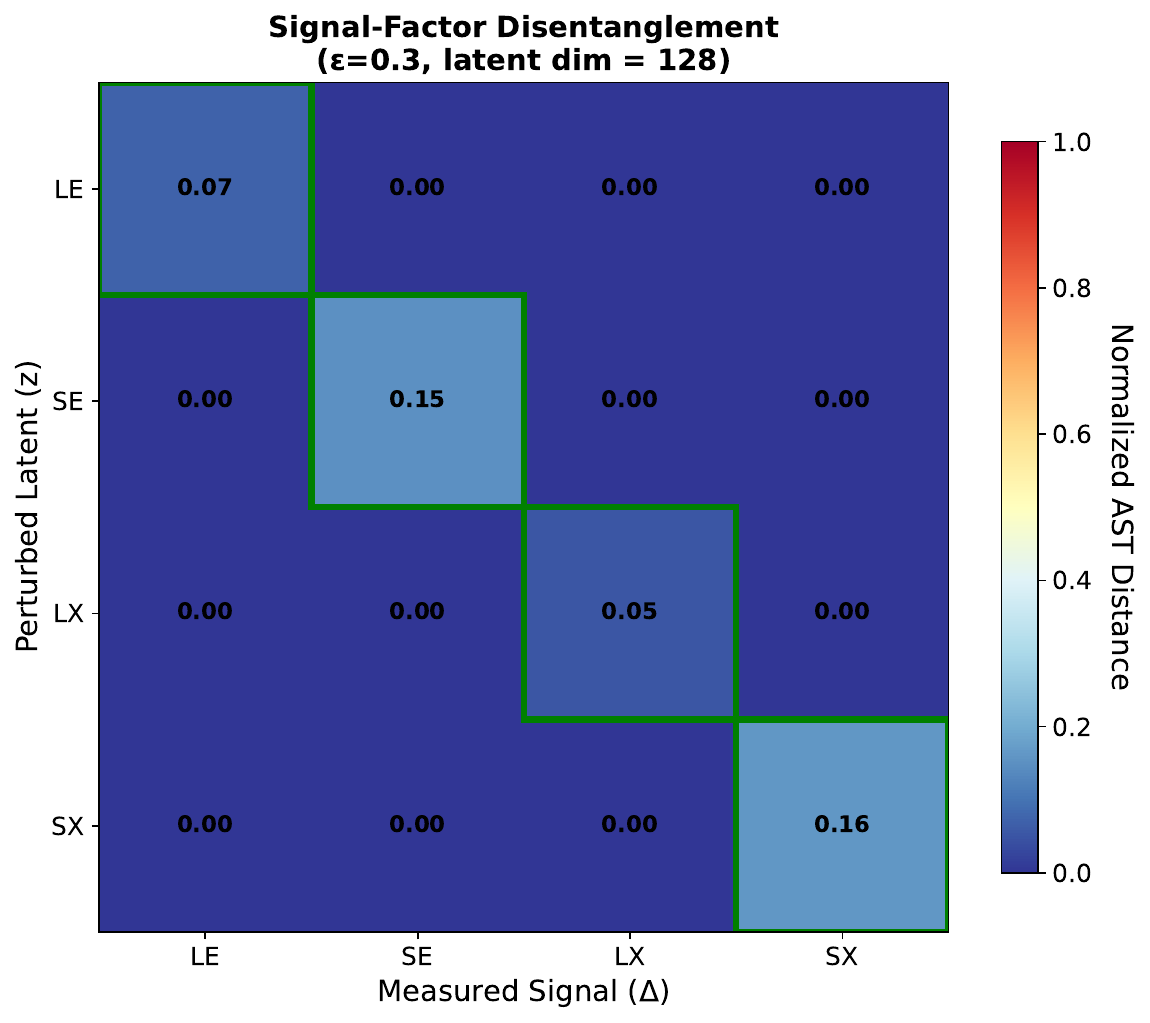}
\caption{Signal-level disentanglement under single-block perturbations ($\epsilon=0.3$, per-signal latent dimension = 32).
Rows indicate which latent block is perturbed.
Columns indicate which program signal changes.
Color shows the normalized token edit distance.
Perturbations affect only the intended signal, with negligible cross-talk.}
\label{fig:signal_disentanglement_heatmap}
\end{figure}

The goal of this test is straightforward. We perturb one latent block and check whether the decoded program changes only in the matching signal expression. For a held-out set of encoded strategies, we use the block-factorized latent representation
\[
z = [z_{le}, z_{se}, z_{lx}, z_{sx}],
\]
corresponding to long entry (LE), short entry (SE), long exit (LX), and short exit (SX).
For each strategy and each latent block $z_k$, we perturb only that block,
\[
z'_k = z_k + \epsilon \eta, \quad \eta \sim \mathcal{N}(0, I),
\]
and keep the other three blocks fixed.
We use $\epsilon = 0.1$.
This value lies within the trust-region regime identified in Section~\ref{sec:behavioral_geometry}.
Each perturbed latent is decoded using greedy decoding.  For each perturbation, we compute the normalized token-level edit distance between the original and perturbed programs separately for each of the four signals. From these per-signal edit distances, we compute two summary metrics. Cross-talk is the fraction of total observed change that appears in \emph{non-target} signals.
The target-only rate is the fraction of perturbations for which all observed changes are confined to the intended signal. Figure~\ref{fig:signal_disentanglement_heatmap} shows the average per-signal edit distance produced by perturbing each latent block.
The result is strongly diagonal.
Perturbing $z_{le}$ changes LE, perturbing $z_{se}$ changes SE, and so on.
Off-diagonal effects are negligible.
Cross-talk is effectively zero, and the target-only rate is $100\%$ across all evaluated signals.

These results show that the learned representation provides reliable component-level control.
Each latent block governs a distinct semantic part of the strategy.

\begin{table*}[t]
\centering
\caption{Walk-forward train, validation, and test folds used for all experiments. All results are reported on out-of-sample test folds with a 10-day embargo between splits to prevent lookahead bias.}
\label{tab:folds}
\begin{tabular}{lccccccccc}
\toprule
Fold & Train Start & Train End & Val Start & Val End & Test Start & Test End & Train Days & Val Days & Test Days \\
\midrule
1 & 2008-01-01 & 2010-06-30 & 2010-06-30 & 2011-01-11 & 2011-01-21 & 2011-07-25 & 910 & 195 & 185 \\
2 & 2011-07-25 & 2014-01-20 & 2014-01-20 & 2014-08-03 & 2014-08-13 & 2015-02-14 & 910 & 195 & 185 \\
3 & 2015-02-14 & 2017-08-12 & 2017-08-12 & 2018-02-23 & 2018-03-05 & 2018-09-06 & 910 & 195 & 185 \\
4 & 2018-09-06 & 2021-03-05 & 2021-03-05 & 2021-09-16 & 2021-09-26 & 2022-03-30 & 910 & 195 & 185 \\
5 & 2022-03-30 & 2024-09-25 & 2024-09-25 & 2025-04-08 & 2025-04-18 & 2025-10-20 & 910 & 195 & 185 \\
\bottomrule
\end{tabular}
\end{table*}

\end{document}